\documentclass[journal,transmag]{IEEEtran}

\usepackage{color}
\usepackage[table,xcdraw]{xcolor}

\usepackage{cite}
\ifCLASSINFOpdf
  \usepackage[pdftex]{graphicx}
\else&
\fi

\usepackage{amsmath}
\usepackage{textcomp}
\usepackage{amssymb}
\usepackage[utf8]{inputenc}
\usepackage[textwidth=1.cm, textsize=tiny]{todonotes}
\usepackage{multirow}
\usepackage{url}

\def \R {\mathbb{R}}
\def \E {\mathbb{E}}
\def \X {\mathcal{X}}
\def \Y {\mathcal{Y}}
\DeclareMathOperator*{\argmin}{arg\,min}

\title{A deep learning architecture for temporal sleep stage classification using multivariate and multimodal time series}

\begin{document}


\author{\IEEEauthorblockN{Stanislas Chambon\IEEEauthorrefmark{1, 2},
Mathieu N. Galtier\IEEEauthorrefmark{2}, 
Pierrick J. Arnal\IEEEauthorrefmark{2},
Gilles Wainrib\IEEEauthorrefmark{3} and
Alexandre Gramfort\IEEEauthorrefmark{1, 4, 5}%,~\IEEEmembership{Fellow,~IEEE}
}
\IEEEauthorblockA{\IEEEauthorrefmark{1}LTCI, Telecom ParisTech, Universit\'e Paris-Saclay, Paris, France}
\IEEEauthorblockA{\IEEEauthorrefmark{2}Research \& Algorithms Team, Rythm inc., Paris, France}
\IEEEauthorblockA{\IEEEauthorrefmark{3}DATA Team, D\'epartement d'Informatique, \'Ecole Normale Sup\'erieure, Paris, France}
%\IEEEauthorblockA{\IEEEauthorrefmark{4}Owkin inc., Paris, France}
\IEEEauthorblockA{\IEEEauthorrefmark{4}INRIA, Universit\'e Paris-Saclay, Paris, France}
\IEEEauthorblockA{\IEEEauthorrefmark{5}CEA, Universit\'e Paris-Saclay, Paris, France}
\thanks{
This work was supported in part by the french Association Nationale de la Recherche et de la Technologie (ANRT) under Grant 2015 / 1005.

S. Chambon is with the Research \& Algorithms Team, Rythm, Paris and Laboratoire Traitement et Communication de l\textquotesingle Information (LTCI), Telecom ParisTech, Université Paris-Saclay, Paris (corresponding author: stanislas@rythm.co).

M. N. Galtier and P. J. Arnal are with the Research \& Algorithms Team, Rythm, Paris (e-mails: mathieu@rythm.co and pierrick@rythm.co).

G. Wainrib is with DATA Team, Département d'Informatique, Ecole Normale Supérieure, Paris (e-mail: gilles.wainrib@ens.fr).

Alexandre Gramfort is with LTCI, Télécom ParisTech, Université Paris-Saclay, Paris and Inria, Université Paris-Saclay, Paris (e-mail: alexandre.gramfort@inria.fr)}}

%\markboth{UNDER REVIEW AS A JOURNAL PAPER AT IEEE TRANSACTION}%
%{correspond authors...}


%\IEEEspecialpapernotice{(In&vited Paper)}

% for Transactions on Magnetics papers, we must declare the abstract and
% index terms PRIOR to the title within the \IEEEtitleabstractindextext
% IEEEtran command as these need to go into the title area created by

\maketitle
\IEEEdisplaynontitleabstractindextext
\IEEEpeerreviewmaketitle

%\IEEEtitleabstractindextext{%
%\stan{new abstract}
\begin{abstract}
\emph{Abstract}:
\color{black}  Sleep stage classification constitutes an important preliminary exam in the diagnosis of sleep disorders. It is traditionally performed by a sleep expert who assigns to each 30\,s of signal a sleep stage, based on the visual inspection of signals such as electroencephalograms (EEG), electrooculograms (EOG), electrocardiograms (ECG) and electromyograms (EMG). We introduce here the first deep learning approach for sleep stage classification that learns end-to-end without computing spectrograms or extracting hand-crafted features, that exploits all multivariate and multimodal Polysomnography (PSG) signals (EEG, EMG and EOG), and that can exploit the temporal context of each 30\,s window of data. For each modality the first layer learns linear spatial filters that exploit the array of sensors to increase the signal-to-noise ratio, and the last layer feeds the learnt representation to a softmax classifier. Our model is compared to alternative automatic approaches based on convolutional networks or decisions trees. Results obtained on 61 publicly available PSG records with up to 20 EEG channels demonstrate that our network architecture yields state-of-the-art performance. Our study reveals a number of insights on the spatio-temporal distribution of the signal of interest: a good trade-off for optimal classification performance measured with balanced accuracy is to use 6 EEG with 2 EOG (left and right) and 3 EMG chin channels. Also exploiting one minute of data before and after each data segment offers the strongest improvement when a limited number of channels is available. As sleep experts, our system exploits the multivariate and multimodal nature of PSG signals in order to deliver state-of-the-art classification performance with a small computational cost.
\end{abstract}

%}

\section{Introduction}
%1 general paragraph about the nature of information
Sleep stage identification, \emph{a.k.a.} \emph{sleep scoring} or \emph{sleep stage classification}, is of great interest to better understand sleep and its disorders. Indeed, the construction of an hypnogram, the sequence of sleep stages over a night, is often involved, as a preliminary exam, in the diagnosis of sleep disorders such as insomnia or sleep apnea~\cite{Berthomier2007a}. Traditionally, this exam is performed as follows. First a subject sleeps with a medical device which performs a polysomnography (PSG), \textit{i.e.}, it records electroencephalography (EEG) signals at different locations over the head, electrooculography (EOG) signals, electromyography (EMG) signals, and eventually more. Second, a human sleep expert looks at the different time series recorded over the night and assigns to each $30$\,s time segment a sleep stage following a reference nomenclature such as American Academy of Sleep Medicine (AASM) rules~\cite{Iber2007} or Rechtschaffen and Kales (RK) rules~\cite{Rechtschaffen1968}. Regarding the AASM rules, $5$ stages are identified: Wake (W), Rapid Eye Movements (REM), Non REM1 (N1), Non REM2 (N2) and Non REM3 (N3) also known as slow wave sleep or even deep sleep. They are characterized by distinct time and frequency patterns and they also differ in proportions over a night. For instance, transitory stages such as N1 are less frequent than REM or N2. In the case of AASM rules, the transitions between two different stages are also documented and the transition rules may modulate the final decision of a human scorer. Indeed, some transitions are prohibited or others are strengthened depending on the occurence of some events such as arousal, K-complex or spindles regarding the transition N1-N2~\cite{Iber2007, Tsinalis2015}. Although very precious information is collected thanks to this exam, sleep scoring is a tedious and time consuming task which is furthermore subject to the scorer subjectivity and variability~\cite{Stephansen2017, Rosenberg2014}.

The use of automatic sleep scoring methods or at least an automatic assistance has been investigated for several years and has driven much interest. From a statistical machine learning perspective, the problem is an imbalanced multi-class prediction problem. State-of-the-art automatic approaches can be classified into two categories depending on whether the features used for classification are extracted using expert knowledge or if they are learnt from the raw signals. Methods of the first category rely on a priori knowledge about the signals and events that enables to design hand-crafted features (see~\cite{Aboalayon2016} for a very extensive list of references).  \color{black}  Methods in the second category consist in learning appropriate feature representations from transformed data~\cite{Vilamala2017, Stephansen2017, Biswal2017, Dong2016} or directly from raw data with convolutional neural networks~\cite{Tsinalis2016, Supratak2017, Sors2017}. Recently, another method was proposed to perform sleep stage classification onto radio waves signals, with an adversarial deep neural network~\cite{Zhao2017}. \color{black}

One of the main statistical learning challenges is the imbalanced nature of the classification task which has important practical implications for this application. Typically sleep stages such as N1 are rare compared to N2 stages. When learning a predictive algorithm with very imbalanced classes, what classically happens is that the resulting system tends to never predict the rarest classes. One way to address this issue is to reweight the model loss function so that the cost of making an error on a rare sample is larger~\cite{He-etal:09}. \color{black}  With an online training approach as used with neural networks, one way to achieve this is to employ \emph{balanced sampling}, \textit{i.e.} to feed the network with batches of data which contain as many data points from each class~\cite{Tsinalis2015, Tsinalis2016, Dong2016, Supratak2017, Sors2017, Biswal2017, Stephansen2017}. \color{black} This indeed prevents the predictive models to be biased towards the most frequent stages. Yet, such a strategy raises the question of the choice of the evaluation metric used. The standard \emph{Accuracy} metric (Acc.) considers that any prediction mistake has the same cost. Imagine that N2 would represent 90\,\% of the data, predicting always N2 would lead to a 90\,\% accuracy, which is obviously bad. A natural way to better evaluate a model in the presence of imbalanced classes is to use the \emph{Balanced Accuracy} (B. Acc.) metric. With this metric the cost of a mistake on a sample of type N2 is inversely proportional to the fraction of samples of type N2 in the data. By doing so, every sleep stage has the same impact on the final figure of merit~\cite{Lajnef2015}.
%\stan{I say it in the last paragraph}
%We will therefore use this metric in our experiment as it was also done in. 

% \ag{make a nice transition with the next paragraph. Don't say too much. You just want to introduce the notion of temporal sleep stage classification}
% \stan{ok}
Another statistical learning challenge concerns the way transition rules are handled. Indeed, as the transition rules may impact the final decision of a scorer, a predictive model might take them into account in order to increase its performance. \color{black}  Doing so is possible by feeding the final classifier with the features from the neighboring time segments~\cite{Stephansen2017, Sors2017, Biswal2017, Supratak2017, Dong2016, Tsinalis2016, Tsinalis2015}. This is referred to as \emph{temporal sleep stage classification}.\color{black}
%\agtodo{this next sentence belongs to discussion}
%stan ok I put it in discussion
%{\color{red} On the other hand, some subjects might exhibit abnormal sleep structures related to a sleep disorder~\cite{Rosenberg2014}. There is thus a trade-off between boosting the classification performance by integrating as much context as possible and not over-fitting sleep transitions in order to not miss a sleep disorder related to a fragmented sleep.} \color{black}

% The transition rules' impact of the classification task has also been addressed by strategies which aim at taking into account the temporal context of the sample to be scored. This is referred to as \emph{temporal sleep stage classification}. Proposed methods consist in feeding a classifier with inputs corresponding to $150$\,s of signal ($60$\,s before and after the sample to classify)~\cite{Tsinalis2015, Tsinalis2016} or feeding a classifier doted of a recurrent network unit with a input sequence of time series or features from neighborging $30$\,s time samples~\cite{Dong2016, Supratak2017}. Such works reportedly improved classification results but they rather focused on using the temporal context to boost classification performances instead of explaining its scope of action.

A number of public sleep datasets contain PSG records with several EEG channels, and additional modalities such as EOG or EMG channels~\cite{OReilly2014}. Although these modalities are used by human experts for sleep scoring, seldom are they considered by automatic systems~\cite{Lajnef2015}.
%\textbf{I have the impression that the next sentence is not so clear.. What do you think ?}
%Focusing only on the EEG modality, it is natural to think that the multivariate nature of EEG data does carry precious information, yet it is often only used to cope with electrode removal or bad channels and not as a leverage to improve the algorithm.
\color{black} 
% \textbf{What about changing it into}
Focusing only on the EEG modality, it is natural to think that the multivariate nature of EEG data does carry precious information. This can be exploited at least to cope with electrode removal or bad channels, and thus improve the robustness of the prediction algorithm. However, this can also be exploited as a leverage to improve the predictive capacities of the algorithm. \color{black}
Indeed, the EEG community has designed a number of methods to increase the signal-to-noise ratio (SNR) of an effect of interest from a full array of sensors. Among these methods are so called  linear spatial filters and include classical techniques such as PCA/ICA~\cite{Parra2005326}, Common Spatial Patterns for BCI applications~\cite{Blankertz2008} or beamforming methods for source localization~\cite{VanVeen1997}. Less classically and more recently various deep learning approaches have been proposed to learn from EEG data~\cite{Mirowski2009, Wulsin2011, Zheng2014, Bashivan2016} and some of these contributions  use a first layer that boils down to a spatial filter ~\cite{Cecotti2008, Cecotti2011, Stober2014, Manor2015, Stober2016, Lawhern2016}. \color{black}  Note that using a deep neural network to learn a feature representation and classify sleep stages on data coming from multiple sensors has been recently investigated in parallel of our work~\cite{Stephansen2017, Biswal2017}. Yet our study further investigates and quantifies how much using a spatial filtering step enhances the prediction performance. \color{black}

%\agtodo{The next paragraph is discussion not introduction. Move or remove if already somewhere.}
%\stan{OK}
%In this paper we adress the sleep stage classification on both a multivariate and temporal angle with a deep learning approach. Our deep net takes into account the multivariate angle by processing several EEG / EOG and EMG channels at the same time.

% %\stan{Could you have a look at this paragraph ?}
% %\color{black} 
% %\ag{We can do better in terms of positioning. We propose a network that can fuse multiple modalities and?}
% In this paper, we introduce an deep learning approach to perform temporal sleep stage classification from multiple modalities (EEG, EOG, EMG) with multivariate inputs (several channels for each modality) in the perspective of maximizing the \emph{Balanced Accuracy}. We then explore our model dependencies with respect to its different parameters.

This paper is organized as follows. First we introduce our end-to-end deep learning approach to perform temporal sleep stage classification using multivariate time series coming from multiple modalities (EEG, EOG, EMG). We furthermore detail how the temporal context of each segment of data can be exploited by our model. Then, we benchmark our approach on publicly available data and compare it to state-of-the-art sleep stage classification methods. Finally, we explore the dependencies of our approach regarding the spatial context, the temporal context and the amount of training data at hand.

\color{black}  
\paragraph*{Notation}
We denote by $X \in \mathbb{R}^{C \times T}$ a segment of multivariate time series with its label $y \in \mathcal{Y}$ which maps to the set $\left\{W, N1, N2, N3, REM \right\}$. Here, $X$ corresponds to a sample lasting 30 seconds and $\mathcal{Y} = \left\{y \in \R^5_+: \sum_{i=1}^5 y_i = 1\right\}$ corresponds to the probability simplex. Precisely, each label is encoded as a vector of $\R^5$ with 4 coefficients equal to $0$ and a single coefficient equal to $1$ which indicates the sleep stage. Here $C$ refers to the number of channels and $T$ to the number of time steps. $\mathcal{S}_t^k = \left\{X_{t-k}, \cdots, X_{t}, \cdots, X_{t+k} \right\}$ stands for an ordered sequence of $2k + 1$ neighboring segments of signal. $\X_k = (\R^{C \times T})^{2k + 1}$ is the space of $2k + 1$ neighboring segments of signal. Finally, $\ell$ stands for the categorical cross entropy loss function. Given a true label $y \in \Y$ and a predicted label $p \in \Y$ it is defined as: $\ell(y, p) = - \sum_{i=1}^5 y_i \log p_i$.

\section{Material and methods}
In this section, we present a deep learning architecture to perform temporal sleep stage classification from multivariate and multimodal time series. We first define formally the classification problem addressed here. Then we present the network architecture used to predict without temporal context ($k=0$). Then we describe the time distributed multivariate network proposed to perform temporal sleep stage classification ($k>0$). Finally, we present and discuss the alternative state-of-the-art methods used for comparison in our experiments.

\subsection{Machine learning problem}
In this paragraph, we formalize in mathematical terms the temporal classification task considered here. Let $k$ be a non-negative integer. Let $f: \X_k \longrightarrow \Y$ stand for a predictive model that belongs to a parametric set denoted $\mathcal{F}$. Here $f$ takes as input an ordered sequence of $2k + 1$ neighboring segments of signal, and outputs a probability vector $p \in \Y$. For simplicity the parameters of the network are not written. The machine learning problem tackled then reads:
\begin{equation}\label{eq:main_problem}
\hat{f} = \argmin_{f \in \mathcal{F}} \E_{x, y \in \X_k \times \Y} \left[ \ell (f(x), y) \right] \enspace .
\end{equation}

Equation \eqref{eq:main_problem} implies that the parameters of the neural network $f$ are optimized by minimizing the expected value of the categorical cross entropy between the output of this network $f(x)$ and the true label $y$.

Whenever $k > 0$ the neural network has access to the temporal context of the segment of signal to classify, it is the \emph{temporal sleep stage classification problem}, and when $k = 0$ the problem boils down to the standard formulation of sleep stage classification.
\color{black}

\subsection{Multivariate Network Architecture}
\begin{figure}[th!]
\centering
\includegraphics[width=0.98\linewidth]{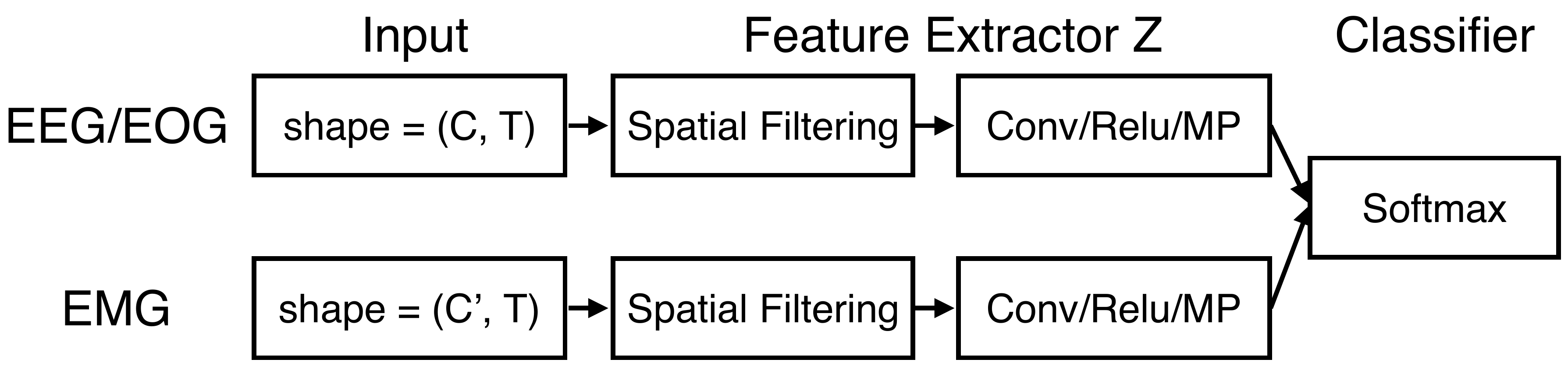} 
\caption{\label{fig:network_architecture}Network general architecture: the network processes $C$ EEG/EOG channels and $C'$ EMG channels through separate pipelines. For each modalitity, it performs spatial filtering and applies convolutions, non linear operations and max pooling (MP) over the time axis. The outputs of the different pipelines are finally concatenated to feed a softmax classifier.
%\ag{Word Convolutions in the box is misleading. If I don't read the caption I assume your network is linear.}
}
\end{figure}

The deep network architecture we propose to perform sleep stage classification from multivariate time series without temporal context ($k=0$) has three key features: linear spatial filtering to estimate so called \emph{virtual channels}, convolutive layers to capture spectral features and separate pipelines for EEG/EOG and EMG respectively. This network constitutes a general feature extractor we denote by $Z: \mathbb{R}^{C \times T} \rightarrow \mathbb{R}^D$, where $D$ is the size of the estimated feature space. Our network can handle various number of input channels and several modalities at the same time. The general architecture is represented in Fig.~\ref{fig:network_architecture}.

We now detail the different blocks of the network, which are summarized in Tab.~\ref{tab:cnn_architecture}. The first layer of the network is a time-independent linear operation
%convolution\ag{don't call this a spatial convolution ! a convolution is a sliding dot product. Here you don't slide over space but apply the same dot product over time. I would call it a time-independent linear operation}, linear operation 
that outputs a set of virtual channels,
each obtained by linear combination of the original input channels. It implements a \emph{spatial filtering} driven by the classification task to perform~\cite{Cecotti2008, Cecotti2011, Stober2014, Manor2015, Stober2016, Lawhern2016}.
%This operation is extremely popular for EEG data processing as it leverages the availability of a full array of sensors to increase the signal-to-noise ratio (SNR) of the effect of interest. This operation commonly referred to as spatial filtering
%\ag{if spatial filtering is mentioned in intro no-need to reintroduce the concept} \stan{ok}
%implies that the network does not directly apply non-linear operations to the original time series, but rather to virtual time series computed as spatial linear combinations of the recorded multivariate time series.
In our experiments, the number of virtual channels was set to the number of input channels making the first layer a multiplication with a square matrix. This square matrix plays the same role as the unmixing matrix estimated by ICA algorithms. This step will be further discussed in the discussion. Note that this first layer based on spatial filters can be implemented with a $2$D valid convolution with kernels of shape $(C, 1)$, see layer 3 in Tab.~\ref{tab:cnn_architecture}.

%\ag{we need refs somewhere on spatial filters and their use on EEG cf. ICA, CSP etc.}.
%XXX: done
Following this linear operation, the dimensions are permuted, see layer 4 in Tab.~\ref{tab:cnn_architecture}. Then two blocks of temporal convolution followed by non-linearity and max pooling are consecutively applied. The parameters have been set for signals sampled at $128$\,Hz. In this case the number of time steps is $T = 128 \times 30 = 3840$. Each block first convolves its input signal with 8 estimated kernels of length 64 with stride $1$ ($\sim 0.5$\,s of record) before applying a rectified linear unit, \emph{a.k.a.} ReLU non-linearity $x \mapsto \mathrm{max}(x, 0)$~\cite{Nair2010}.
The outputs are then reduced along the time axis with a max pooling layer (size of $16$ without overlap). The output of the two convolution blocks is finally passed through a dropout layer~\cite{Srivastava2014} which randomly prevents updates of $25\%$ of its output neurons at each gradient step.
 % Although the architecture was developped for a $128$\,Hz sampling frequency, it can be easily adapted to any other one.

As represented in Fig.~\ref{fig:network_architecture}, we process jointly the EEG and EOG time series since these modalities are comparable in magnitudes and both measure similar signals, namely electric potential up to a few hundreds of microvolts on the surface of the scalp. The same idea is used by EEG practitioners when the EOG channels are kept in the ICA decomposition to better reject EOG artifacts~\cite{Joyce2004}. The EMG time series which have different statistical and spectral properties are processed in a parallel pipeline.

The resulting outputs are then concatenated to form the feature space of dimension $D$ before being fed into a final layer with $5$ neurons and a \emph{softmax} non-linearity to obtain a probability vector which sums to one. This final layer is referred to as a \emph{softmax classifier}~\cite{Goodfellow-et-al-2016}. \color{black}   Let $a \in \R^5$ be the pre-activation of the last layer. The output of the network is a vector $p \in \Y$. $p$ is obtained as: $p_i = \exp(a_i) / \sum_{j=1}^5 \exp(a_j)$.
\color{black}.

\begin{table*}[ht!]
\centering
\def\arraystretch{1.5}
\begin{tabular}{l|lllllllll}
\textbf{}                                                                                               & \textbf{Layer} & \textbf{Layer Type} & \textbf{\# filters} & \textbf{\# params}    & \textbf{size} & \textbf{stride} & \textbf{Output dimension} & \textbf{Activation} & \textbf{Mode} \\ \hline
\rowcolor[HTML]{EFEFEF} 
\cellcolor[HTML]{EFEFEF}                                                                                & 1              & Input               &                     &                       &               &                 & (C, T)                    &                     &               \\
\rowcolor[HTML]{EFEFEF} 
\cellcolor[HTML]{EFEFEF}                                                                                & 2              & Reshape             &                     &                       &               &                 & (C, T, 1)                 &                     &               \\
\rowcolor[HTML]{EFEFEF} 
\cellcolor[HTML]{EFEFEF}                                                                                & 3              & Convolution 2D      & C                   & C * C                 & (C, 1)        & (1, 1)          & (1, T, C)                 & Linear              &               \\
\rowcolor[HTML]{EFEFEF} 
\cellcolor[HTML]{EFEFEF}                                                                                & 4              & Permute             &                     &                       &               &                 & (C, T, 1)                 &                     &               \\
\rowcolor[HTML]{EFEFEF} 
\cellcolor[HTML]{EFEFEF}                                                                                & 5              & convolution 2D      & 8                   & 8 * 64 + 8            & (1, 64)       & (1, 1)          & (C, T, 8)                 & Relu                & same          \\
\rowcolor[HTML]{EFEFEF} 
\cellcolor[HTML]{EFEFEF}                                                                                & 6              & maxpooling 2D       &                     &                       & (1, 16)       & (1, 16)         & (C, T // 16, 8)           &                     &               \\
\rowcolor[HTML]{EFEFEF} 
\cellcolor[HTML]{EFEFEF}                                                                                & 7              & convolution 2D      & 8                   & 8 * 8 * 64 + 8        & (1, 64)       & (1, 1)          & (C, T // 16, 8)           & Relu                & same          \\
\rowcolor[HTML]{EFEFEF} 
\cellcolor[HTML]{EFEFEF}                                                                                & 8              & maxpooling 2D       &                     &                       & (1, 16)       & (1, 16)         & (C, T // 256, 8)           &                     &               \\
\rowcolor[HTML]{EFEFEF} 
\cellcolor[HTML]{EFEFEF}                                                                                & 9              & Flatten             &                     &                       &               &                 & (C * (T // 256) * 8)       &                     &               \\
\rowcolor[HTML]{EFEFEF} 
\multirow{-10}{*}{\cellcolor[HTML]{EFEFEF}\begin{tabular}[c]{@{}l@{}}Features\\ Extractor\end{tabular}} & 10             & Dropout (50\%)      &                     &                       &               &                 & (C * (T // 256) * 8)       &                     &               \\
\rowcolor[HTML]{D8D8D8} 
Classifier                                                                                              & 11             & Dense               &                     & 5 * (C * T // 256 * 8) &               &                 & 5                         & Softmax             &              
\end{tabular}
\vspace{3mm}
\caption{Detailed architecture for the feature extractor for $C$ EEG channels with time series of length $T$. The same architecture is employed for $C'$ EMG channels. When both EEG / EOG and EMG are considered, the outputs of the dropout layers are concatenated and fed into the final classifier. The number of parameters of the final dense layer becomes thus equal to $5 \times ((C + C') \times (T$ // $256) \times 8$). \label{tab:cnn_architecture}}
\end{table*}

\subsection{Time Distributed Multivariate Network}
In this paragraph, we describe the \emph{Time Distributed Multivariate Network} we propose to perform \emph{temporal sleep stage classification} ($k > 0$). It builds on the \emph{Multivariate Network Architecture} presented previously and distributes it in time to take into account the temporal context. Indeed a sample of class N2 is very likely to be close to another N2 sample, but also to an N1 or an N3 sample~\cite{Iber2007}.

\begin{figure}[ht!]
\centering
\includegraphics[width=0.98\linewidth]{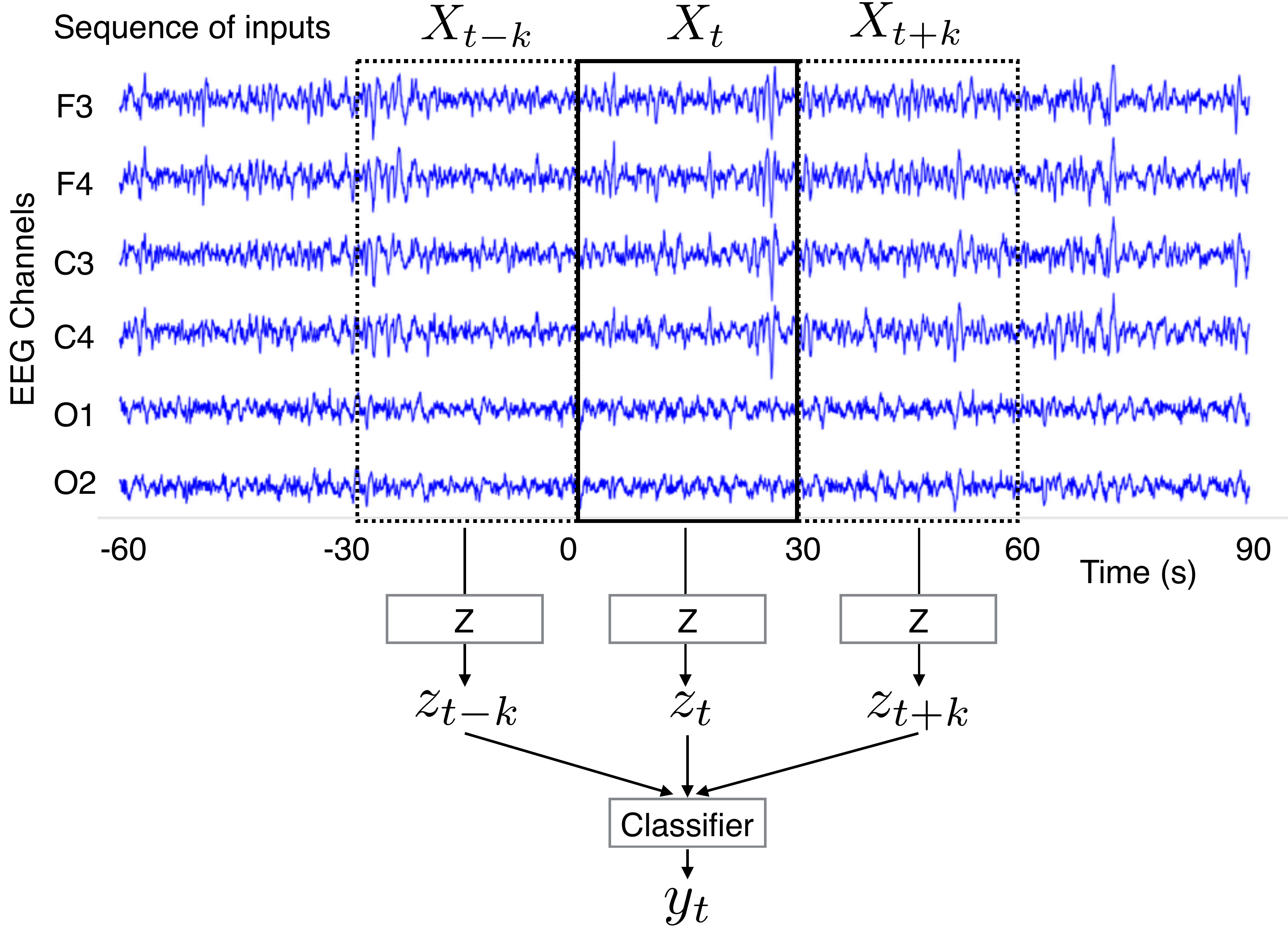} 
\caption{\label{fig:time_distributed_model}Time distributed architecture to process a sequence of inputs $\mathcal{S}_t^k = \left\{X_{t-k}, \cdots, X_t, \cdots, X_{t+k} \right\}$ with $k=1$. $X_k$ stands for the multivariate input data over $30$\,s that is fed into the feature extractor $Z$. Features are extracted from consecutive $30$\,s samples: $X_{t-k}, \cdots, X_t, \cdots, X_{t+k} $. Then the obtained features are aggregated $\left[z_{t-k}, \cdots, z_t, \cdots, z_{t+k}\right]$. The resulting aggregation of features is finally fed into a classifier to predict the label $y_t$ associated to the sample $X_t$.}
\end{figure}

To take into account the statistical properties of the signals before and after the sample of interest, we propose to aggregate the different features extracted by $Z$ on a number of time segments preceding or following the sample of interest. More formally, let $\mathcal{S}_t^k = \left\{X_{t - k}, \cdots, X_t, \cdots, X_{t+k} \right\} \in \X_k$ be a sequence of $2k + 1$ neighboring samples ($k$ samples in the past and $k$ samples in the future). Distributing in time the features extractor consists in applying $Z$ to each sample in $\mathcal{S}_t^k$ and aggregating the $2k+1$ outputs forming a vector of size $D(2k+1)$. Then, the obtained vector is fed into the final softmax classifier. This is summarized in Fig.~\ref{fig:time_distributed_model}.

\subsection{Training}
%\ag{I am wondering if the next paragraph should not be in a dedicated subsection. See how other papers have done it.}
%\stan{ok, it makes sense}
% Training was performed by minimizing the categorical cross entropy, between the true label $y \in \Y$ of a sample $x \in \X$ and the predicted value $p = \hat{f}(x)$ where $\hat{f}$ stands for the neural network.

%Let $x, y \in \mathcal{X} \times \mathcal{Y}$ a sample and its associated label encoded in $\left\{0, 1\right\}^5$. Let $p \in \left[0, 1\right]^5$ be the output of the neural network. The cross entropy between $y$ and $p$ reads:
%\begin{equation}\label{eq:cross_entropy}
%h = - \sum_{i=1}^5 y_i \log p_i
%\end{equation}
%\color{black}

The minimization in \eqref{eq:main_problem} is done with an online procedure based on stochastic gradient descent using mini batches of data. Yet, to be able to learn to discriminate under-represented classes (typically W and N1 stages), and since we are interested in optimizing the balanced accuracy, %\ag{did you say this before? ça tombe comme un cheveu sur la soupe non? this diserves as mathieu pointed out a full paragraph of justification in the introduction}
%\stan{ok}
we propose to balance the distribution of each class in minibatches of size 128. As we have 5 classes it means that during training, each batch has about 20\% of samples of each class. The \emph{Adam} optimizer~\cite{KingmaB14} is used for optimization with the following parameters $\mathrm{\alpha}=0.001$ (learning rate), $\beta_1=0.9$, $\beta_2=0.999$ and $\epsilon=10^{-8}$.
 % and $\mathrm{decay}=0$.
%\ag{which one?}.
%XXX: Keras ones, given above

An early stopping callback on the validation loss with patience of 5 epochs was used to stop the training process when no improvements were detected. Weights were initialized with a normal distribution with mean $\mu = 0$, and standard deviation $\sigma = 0.1$. Those values were obtained empirically by monitoring the loss during training. The implementation was written in \emph{Keras}~\cite{chollet2015keras} with a \emph{Tensorflow} backend~\cite{tensorflow2015}.
% \ag{is this initialization standard? did you see it recommended somewhere?}.
% \stan{not standard, I carried out experiment on it months ago to get the best initialization parameters for the different standardization techniques used, should I redo them ?}

The training of the time distributed network was done in two steps. First, we trained the multivariate network, especially its feature extractor part $Z_t$ without temporal context ($k=0$). The trained model was then used to set the weights of the feature extractor distributed in time. Second, we freezed the weights of the feature extractor distributed in time and we trained the final softmax classifier with aggregated features.

\section{Experiments}
\label{sec:4_experiments}
In this section, we first introduce the dataset and the preprocessing steps used. Then, we present the different features extractors of the literature which we use in our benchmark. We then present the experiments which aim at (i) establishing a general benchmark of our feature extractor against state-of-the art approaches in univariate (single derivation) and bivariate (2 channels) contexts, (ii) studying the influence of the spatial context, (iii) evaluating the gain obtained by using the temporal context and (iv) evaluating the impact of the quantity of training data.

\subsection{Data and preprocessing steps}
Data used in our experiments is the publicly available MASS dataset - session 3~\cite{OReilly2014}. It corresponds to $62$ night records, each one coming from a different subject. Because of preprocessing issues we removed the record \emph{01-03-0034}. Each record contains data from 20 EEG channels which were referenced with respect to the $A2$ electrode. We did not modify the referencing scheme, hence removed the $A2$ electrode from our study. \color{black}  Each record also includes signals from $2$ EOG (horizontal left and right) and $3$ EMG channels (chin channels) that we considered as additional modalities.\color{black}

\color{black}  
The time series from all the available sensors were first low-pass filtered with a $30$\,Hz cutoff frequency. Then they were downsampled to a sampling rate of $128$\,Hz. The downsampling step speeds up the computations for the neural networks, while keeping the information up to $64$\,Hz (Nyquist frequency). Downsampling and low / band pass filtering are commonly used preprocessing steps~\cite{Stephansen2017, Lajnef2015}. \color{black} The data extraction and the filtering steps were performed with the \emph{MNE software}~\cite{Gramfort2014}. The filter employed was a zero-phase finite impulse response (FIR) filter with transition bandwidth of approximately $7$\,Hz. Sleep stages were marked according to the AASM rules by a single sleep expert per record~\cite{OReilly2014, Iber2007}. \color{black}   When investigating the use of temporal context by feeding the predictors with sequences of consecutive samples $S_k$, we used zero padding to complete the samples at the beginning and at the end of the night. This enables to feed the models with all the samples of a night record while keeping fixed the dimension of the input batches. \color{black}

The time series fed into the different neural networks were additionnaly standardized. Indeed, for each channel, every $30$\,s sample is standardized individually such that it has zero mean and unit variance.
%\stan{the standardization is done channel by channel, sample by sample. Actually more and more I think about removing the division by the std and keep only the centering operation as done in image classification}
%\ag{this means that xgboost has info that the network does not have, but maybe the mean and variance have no predictive power... also it is unclear. Does it mean np.std(np.ravel(X)) == 1 and np.mean(np.ravel(X)) == 0 or is it channel by channel?}
For the specific task of sleep stage classification this is particularly relevant since records are carried out over nearly 8 hours. During such a long period the recording conditions vary such as skin humidity, body temperature, body movements or even worse electrode contact loss. Giving to each $30$\,s time series the same first and second order moments enables to cope with this likely covariate shift that may occur during a night record. This operation only rescales the frequency powers in every frequency band, without altering their relative amplitudes where the discriminant information for the considered sleep stage classification task lies (see Parseval's theorem).
%\ag{with your experiments you can do better than supposing}
%\stan{ok I do that}
Note that this preprocessing step can be done online before feeding the network with a batch of data.

\color{black}  
Cross-validation was used to have an unbiased estimate of the performance of our model on unseen records. To reduce variance in the reported scores, the data were randomly split 5 times between train, validation and testing set. The splits were performed with respect to records in order to guarantee that a record used in the training set was never used in the validation or the testing set. For each split, 41 records were included in the training set, 10 records in the validation set and 10 records in the testing set.
\color{black}

%\stan{ok}
%\ag{this next paragraph should not be there. You will have explained Balanced Accuracy in intro and you can explain CM when you show them for the first time.}
%To measure the performance of the different algorithms, three metrics were employed: the \emph{Accuracy} (\emph{Acc.}) and the \emph{Balanced Accuracy} (\emph{B. Acc.}) and \emph{Confusion Matrices} (\emph{C.M.}). These metrics were computed on each testing record individually since what matters is the performance of an algorithm on a given nigth record. As regards the \emph{C.M.} presented in the paper, they were obtained by (i) computing a total \emph{C.M.} as the sum of the \emph{C.M.} evaluated per testing subject (ii) normalizing each row of the total \emph{C.M.} such that it sums up to $1$.

\color{black}  
\subsection{Related work and compared approaches}
We now introduce the three state-of-the-art approaches that we used for comparison with our approach: a gradient boosting classifier~\cite{Friedman00greedyfunction} trained on hand-crafted features and two convolutional networks trained on raw univariate time series following the approach of~\cite{Tsinalis2016} and \cite{Supratak2017}. %, \cite{Sors2017}, \cite{Stephansen2017}.

\subsubsection{Features based approach}
The \emph{Gradient Boosting} model was learnt on hand-crafted features: time domain features and frequency domain features computed for each input sensor as described in~\cite{Lajnef2015}. More precisely, we extracted from each channel the power and relative power in 5 bands: $\delta$ ($0.5 - 4.5 $ \,Hz), $\theta$ ($4.5 - 8.5$ \,Hz), $\alpha$ ($8.5-11.5$\,Hz), $\sigma$ ($11.5 - 15.5$\,Hz), $\beta$ ($15.5 - 30$\,Hz), giving both $5$ features. We furthermore extracted power ratios between these bands (which amount for $5\times4/2=10$ supplementary features) and spectral entropy features as well as statistics such as mean, variance, skewness, kurtosis, $75\%$ quantile. This gives in the end a set of $26$ features per channel.
%\agtodo{What do you mean by relative power? power in band divied by full power? if you standardize channels it's the same information as full power is 1.}
% XXX: good remakr. but there is no problem on this point: we say that we standardize the input time series only for neural networks not for the xgb.

%\stan{yes, you're right, but I just kept all the time and frequency features Lajnef was extracting in his paper. But he was using trees of svm.}
%\ag{does not make sense to extract standard deviation and variance with a GBRT}
%\ag{give more details on features or at least how many you have per channel}
%\stan{ok, done}
The implementation used is from the \emph{XGBoost} package~\cite{Chen}, which internally employs decisions trees. This model is known for its high predictive performance, robustness to outliers, robustness to unbalanced classes and parallel search of the best split. Training was performed by minimizing also the categorical cross entropy. The training set was balanced using under sampling. The maximum number of trees in the model was set to $1000$. An early stopping callback on the validation categorical cross entropy with patience equal to $10$ was used to stop the training when no improvement was observed. Training never led to more than $1000$ trees in a model.
% \color{black}  Stan: integrate results on balanced sampling. speak about it here. \color{black}.

The model has several hyper-parameters that need to be tuned to improve classification performances and cope with unbalanced classes. To find the best hyper-parameters for each experiment, we performed random searches with the \emph{hyperopt} Python package~\cite{Bergstra2013}. Concretely, we considered only the data from the training and validation subjects at hand. For each set of hyper-parameters, we trained and evaluated the classifier on data from $5$ different splits of training and evaluation subjects ($80\%$ for training $20\%$ for evaluation). The search was done with $50$ sets of hyper-parameters and the set which achieved the best balanced accuracy averaged on the $5$ splits was selected.
%\ag{how did you do the cross-validation? what fraction for train and test?}
%\stan{ok adressed}
 The following parameters were tuned: learning rate in interval $\left[10^{-4}, 10^{-1}\right]$, the minimum weight of a child tree in set $\left\{1, 2, \cdots, 10 \right\}$, the maximum depth of trees in $\left\{1, 2, \cdots, 10 \right\}$, the regularization parameter in $\left[0, 1\right]$, the subsampling parameter in $\left[0.5, 1\right]$, the sampling level of columns by tree in $\left[0.5, 1\right]$.
%\ag{I removed all the python variables. It's .tex not code.}
%\stan{ok you're right}

\subsubsection{Convolutional networks on raw univariate time series}
We reimplemented and benchmarked 2 end-to-end deep learning approaches. We detail each of them in the following paragraphs and explain how we used these methods.

\paragraph{Tsinalis et al. 2016}
The approach by \emph{Tsinalis et al. 2016}~\cite{Tsinalis2016} is a deep convolution network that processes univariate time series (a single EEG signal). It was reimplemented according to the paper details. The approach originally takes into account the temporal context, by feeding the network with $150$\,s of signals, \textit{i.e.} the sample to classify plus the $2$ previous and $2$ following samples. When used without temporal context in the experiments, the network is fed with $30$\,s samples.

Training was performed by minimizing the categorical cross entropy, and a similar balanced sampling strategy with \emph{Adam} optimizer was used. An additional $\ell_2$ regularization set to $0.01$ was applied onto the convolution filters~\cite{Tsinalis2016}. The code was written in \emph{Keras}~\cite{chollet2015keras} with a \emph{Tensorflow} backend~\cite{tensorflow2015}.

\paragraph{Supratak et al. 2017}
The approach by \emph{Supratak et al. 2017}~\cite{Supratak2017} is also an end-to-end deep convolutional network which contains two blocks: a feature extractor that processes the frequency content of the signal and a recurrent neural network that processes a sequence of consecutive $30$\,s samples of signal. The feature extractor processes low frequency information and high frequency information into two distinct convolutional sub-neural networks before merging the feature representations. The resulting tensor is then fed into a softmax classifier. This block is trained with balanced sampling. Then the feature extractor is linked to a recurrent neural network composed of 2 bi-LSTM layers. The whole architecture is fed with sequences of $25$ consecutive $30$\,s samples from the same record.% After one record has been processed, the state of the LSTM cells are reset.

The first block was used for comparison in our experiment. Its training was performed by minimizing the categorical crossentropy, and a balanced sampling strategy with \emph{Adam} optimizer was used. The code was written in \emph{Keras}~\cite{chollet2015keras} with a \emph{Tensorflow} backend~\cite{tensorflow2015}.
\color{black}

\color{black}  
\subsection{Experiment 1: Comparison of feature extractors on the Fz / Cz channels}
In this experiment, we perform a general benchmark of our feature extractor against hand-crafted features classified with \emph{Gradient Boosting}, and the two network architectures just described~\cite{Tsinalis2016,Supratak2017}. The purpose of this experiment is to benchmark different feature representations on a similar spatial context, Fz-Cz, without using the temporal context, and to emphasize the benefits of processing multivariate time series instead of a pre-computed derivation.

Only time series coming from the channels Fz and Cz are considered here. First, the four predictive models were fed with the time series or the features from the derivation Fz-Cz that was computed manually. Second, our approach was fed with the time series from the derivations Fz-A2 and Cz-A2, \textit{i.e.}, the original time series of the dataset with pre-computed references. This version of our approach is referred to as \emph{Proposed approach - multivariate}.
No temporal context was used for this experiment ($k=0$).

Finally, the experiment was carried out using balanced sampling at training time. For \emph{Gradient Boosting}, an under sampling strategy was used to balance the training and the validation sets.

\begin{figure}[ht!]
\centering
\includegraphics[width=0.98\linewidth]{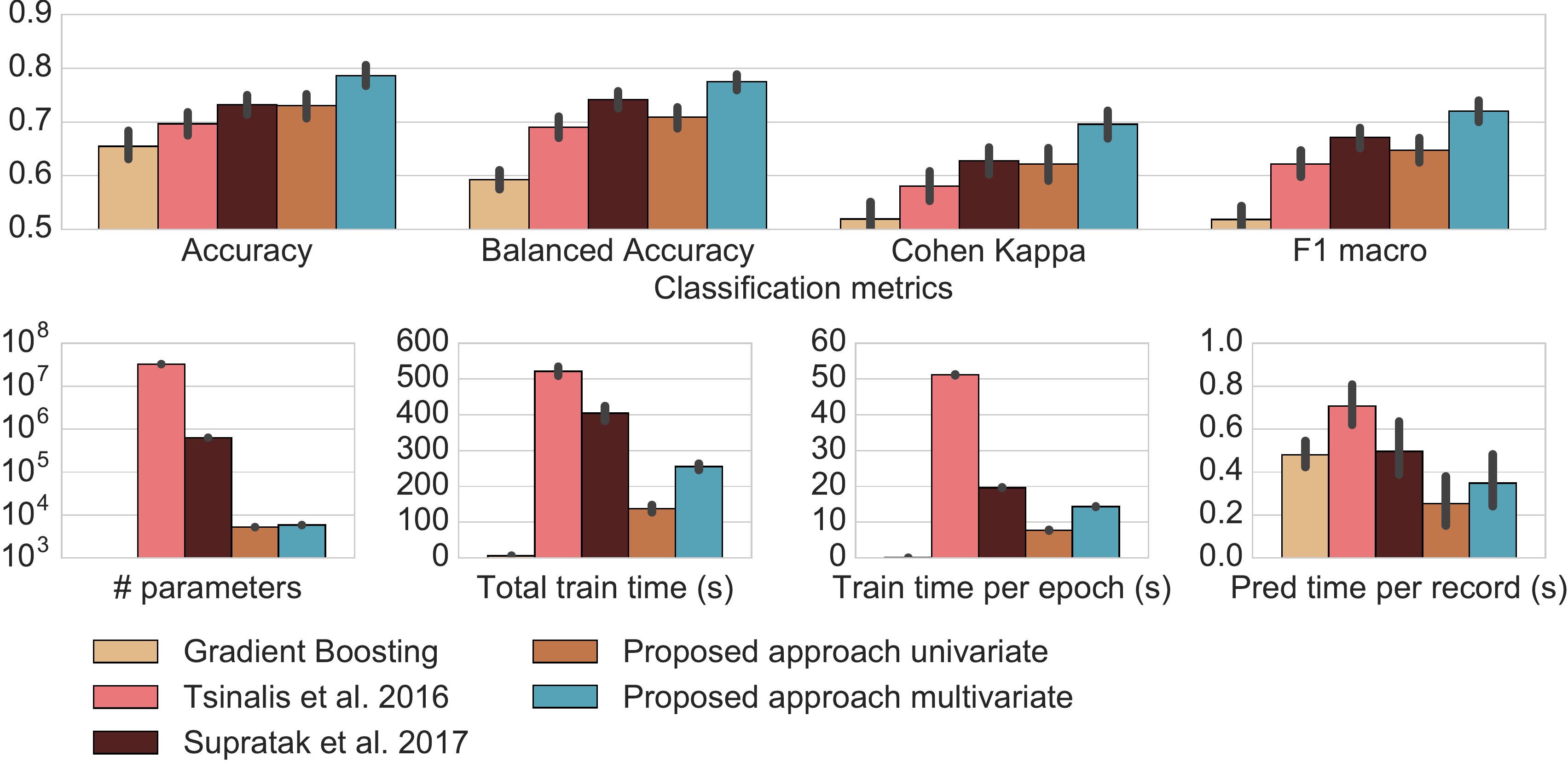}
\caption{\label{fig:benchmark_general_metrics}General classification and run time metrics of several feature extractors benchmarked on the Fz-Cz derivation or Fz-A2, Cz-A2 channels. The proposed approach trained on Fz-A2, Cz-A2 channels obtained higher classification performance than the other feature extractor trained on the Fz-Cz derivation, included its univariate counted-part while having a very low number of parameters and run time at training and prediction time.}
\end{figure}

\begin{figure*}[ht!]
\centering
\includegraphics[width=0.98\linewidth]{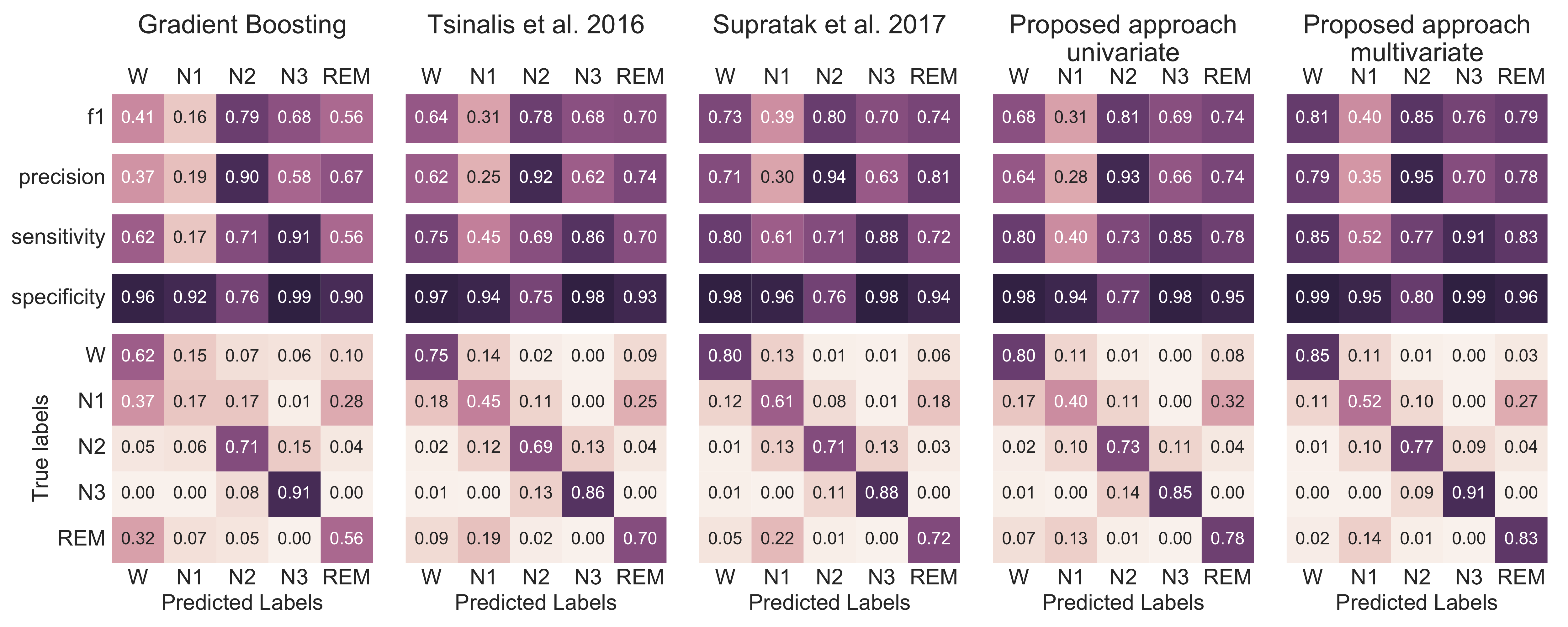}
\caption{\label{fig:benchmark_per_class_metrics}Per class metrics of several feature extractor trained on the Fz-Cz derivation or Fz-A2, Cz-A2 channels.}
\end{figure*}

The performance of the different algorithms is evaluated with general classification metrics: \emph{Accuracy}, \emph{Balanced Accuracy}, \emph{Cohen Kappa}, \emph{F1 score}. Furthermore, run time metrics were computed such as: the \emph{number of parameters}, the \emph{total training time}, the \emph{training time per pass over the train set} (called epoch), the \emph{prediction time per record} (nearly 1k samples). These metrics are reported in Fig.~\ref{fig:benchmark_general_metrics}. Finally per class metrics were used: \emph{F1}, \emph{Precision}, \emph{Sensitivity}, \emph{Specificity} along with confusion matrices (\emph{C.M.}). The \emph{C.M.} were obtained by (i) normalizing the \emph{C.M.} evaluated per testing subject such that its rows sum up to $1$, (ii) computing the average \emph{C.M.} over all testing subjects. These metrics are reported in Fig.~\ref{fig:benchmark_per_class_metrics}

It can be observed in Fig.~\ref{fig:benchmark_general_metrics} that our feature extractor reaches classification performance comparable to that obtained by \emph{Supratak et al. 2017} and higher than those from \emph{Tsinalis et al. 2016} and \emph{Gradient Boosting} on the Fz-Cz derivation. It also uses a very low number of parameters and a low training and prediction run time compared to the other deep learning approaches.

Furthermore, the proposed feature extractor trained on the Fz-A2, Cz-A2 channels, \emph{i.e.} that is fed with multivariate time series, significantly outperforms its univariate counterpart and the other feature extractors which receive univariate time series. Processing two channels instead of a single induces a limited increase in number of parameters, training and prediction run time.

Besides, in Fig.~\ref{fig:benchmark_per_class_metrics}., the univariate proposed method, trained on Fz-Cz, yields equal or higher diagonal coefficients in its confusion matrix than the other feature extractors for sleep stages W, N1, N3. \emph{Supratak et al. 2017} outperforms the different univariate approaches on N1 and N3.

Moreover, the multivariate proposed approach yields higher diagonal coefficients in its confusion matrix than its univariate counterpart and the other feature extractor, except for N1 where \emph{Supratak et al. 2017} exhibits the highest classification accuracy.
The analysis of the other per-class metrics agree with these facts.

\color{black}

\subsection{Experiment 2: More sensors increase performance}
\begin{figure}[ht!]
\centering
\includegraphics[width=0.96\linewidth]{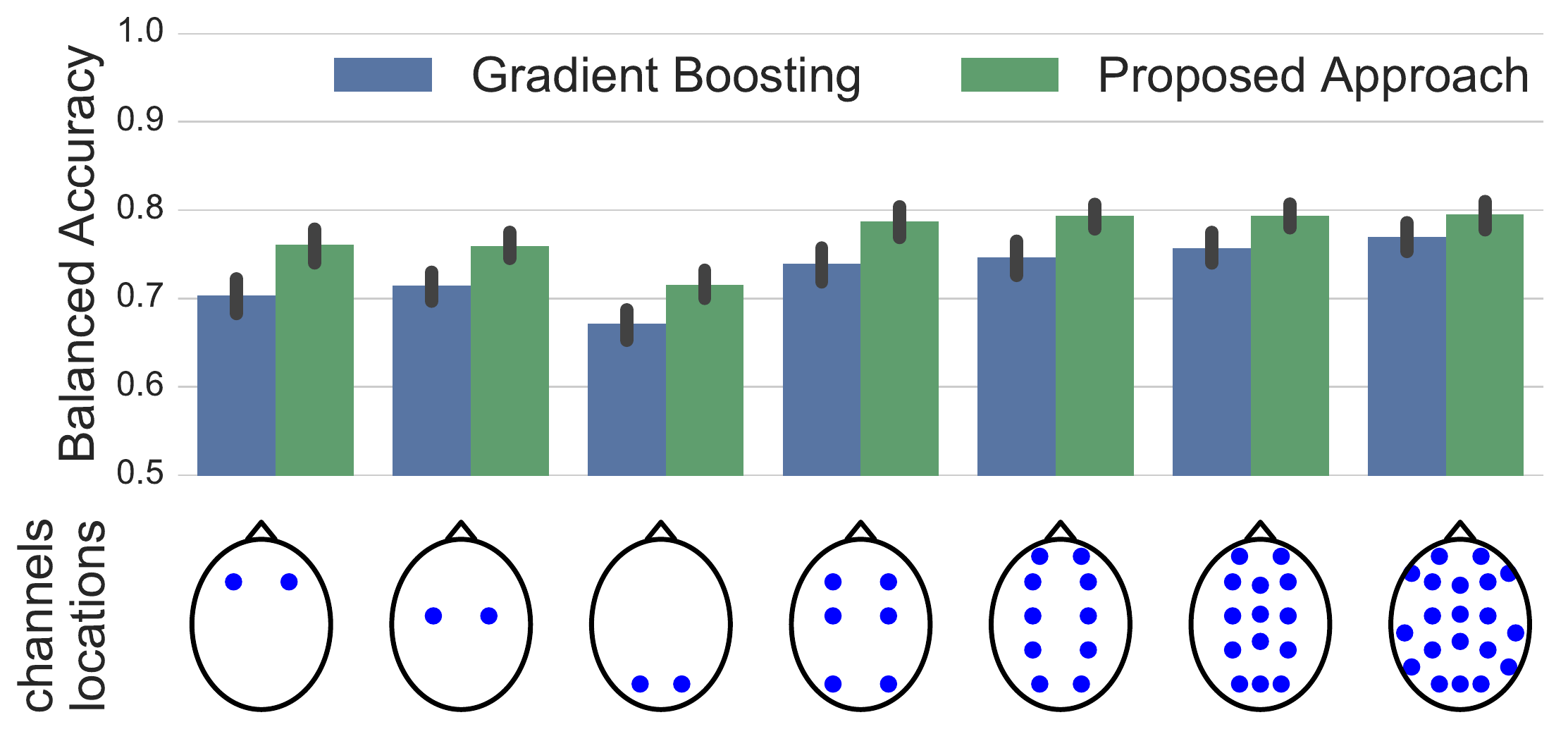}
\caption{\label{fig:influence_spatial_configuration}Influence of channel selection on the classification performances: increasing the number of EEG sensors increases B. Acc.}
\end{figure}

In this experiment, we investigated the influence of the multivariate spatial context on the performance of our approach. 
% 
% However the gain becomes relatively small when more than $6$ EEG channels are taken into account. Adding the EMG and EOG modalities gives an additional gain of performance.
% 
% \subsubsection{Setup}
We considered $7$ different configurations of EEG sensors which varied both in the number of recording sensors from $2$ to $20$ as well as in their positions over the head. We report the classification results for each configuration in Fig.~\ref{fig:influence_spatial_configuration}.

% \subsubsection{Varying the EEG input channes}
One observes that both \emph{Gradient Boosting} and our approach benefit from the increased number of EEG sensors. However, the \emph{B. Acc.} obtained with our approach does not improve once we have $6$ well distributed channels. This is certainly due to the redundancy of the EEG channels, yet more channels could make on some data the model more robust to the presence of bad sensors. First, this demonstrates that it is worth adding more EEG sensors, but up to a certain point. Second, it shows that our approach exploits well the multivariate nature of signals to improve classification performances. Third, it shows that the channel agnostic features extractor, \textit{i.e.} the use of the spatial projection and the features extractor is a good option to fully exploit the spatial distribution of the sensors.

\begin{figure}[ht!]
\centering
\includegraphics[width=0.96\linewidth]{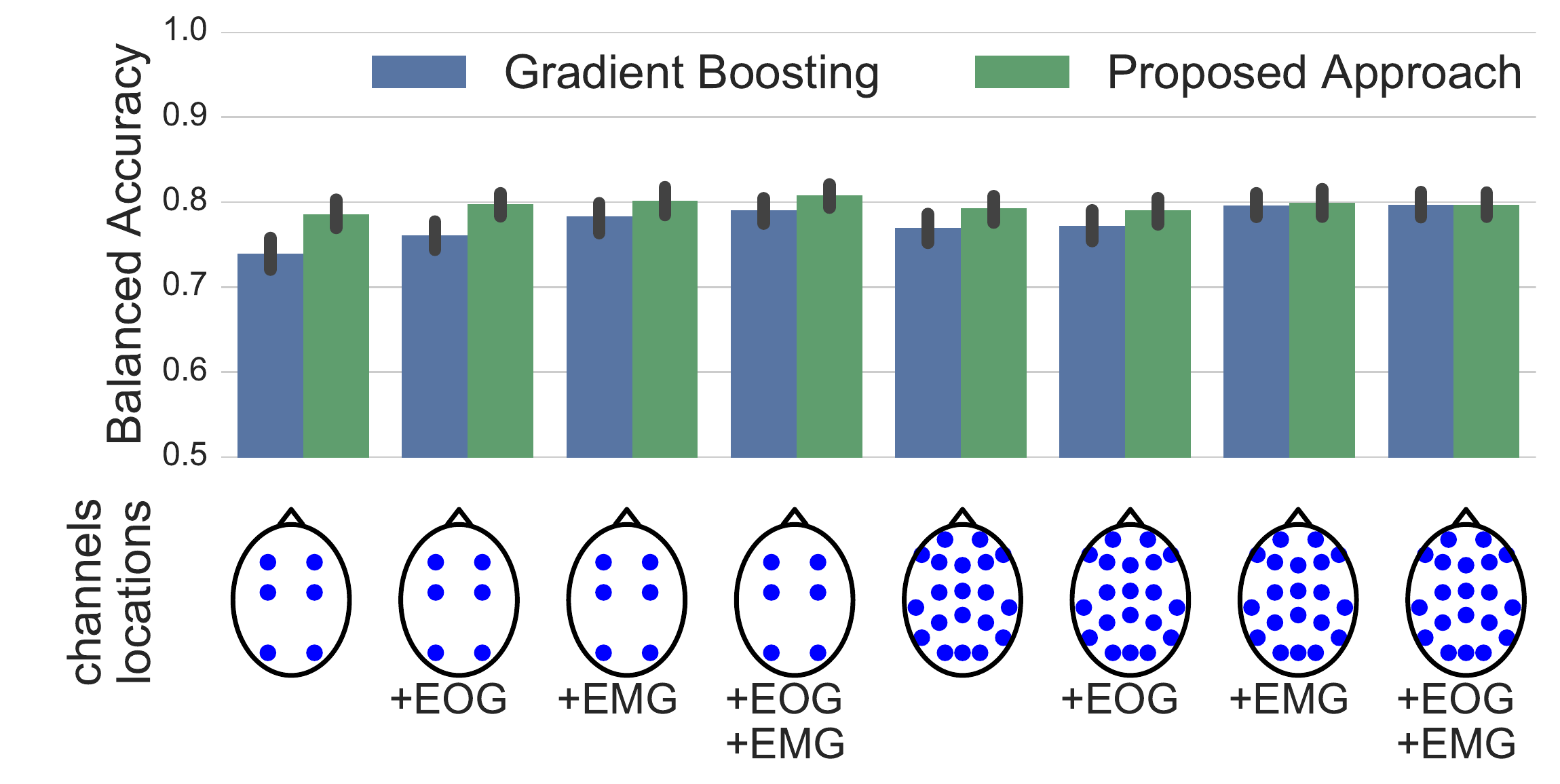}
\caption{\label{fig:influence_additional_modalities}Influence of additional modalities on the classification performances: adding EOG and EMG induces a boost in performance}
\end{figure}

Restricting the number of EEG channels to $6$ and $20$, we further investigated the influence of additionnal modalities (EOG, EMG). Classification results are provided in Fig.~\ref{fig:influence_additional_modalities}.

% \subsubsection{Considering additional modalities}
Considering additional modalities also increases the classification performances of the considered classifiers. It gives them a significative boost of performance, especially when the EMG modality is considered. This means that both approaches successfully integrate the new features with the previous ones.
% More importantly in our perspective, our approach can extract relevant information from both modalities (EEG and EMG) and merge it efficiently to the one already at hand.
This suggests that our feature extractor was sufficiently data agnostic and versatile to handle both modalities. Finally, it again stresses the importance of considering the spatial context, here the additionnal modalities, to improve classification performances.

%Interestingly adding the EOG time series does not improve significantly the performances of the approach compared to adding the EMG ones. This might be explained by the fact that the information carried by EOG channels is already present in the frontal EEG electrodes.

Interestingly, the boost of performance is more important in the $6$ channel setting rather than in the $20$ channel setting. We further observe that both EEG configuration with EOG and EMG modalities reach the same performances. Thus, the use of additional modalities compensate the use of a larger spatial context in this situation. Practically speaking, to obtain the highest performances at a reduced computational cost, one shall consider few well located EEG sensors with additional modalities.

\subsection{Experiment 3: Temporal context boosts performance}
In this experiment, we investigate the influence of the temporal context on the classification performances and demonstrate that considering the data from the neighboring samples increases classification performances especially if the spatial context is limited. \color{black} 
We also report what is the impact of temporal context on confusion matrices, and also on the matrices of transition probabilities between sleep stages. The coefficient $P_{ij}$ of the transition matrix $P \in \R^{5 \time 5}$ is equal to the probability of going from a sleep stage $i$ to a sleep stage $j$.\color{black}

\begin{figure}[ht!]
\centering
\includegraphics[width=0.98\linewidth]{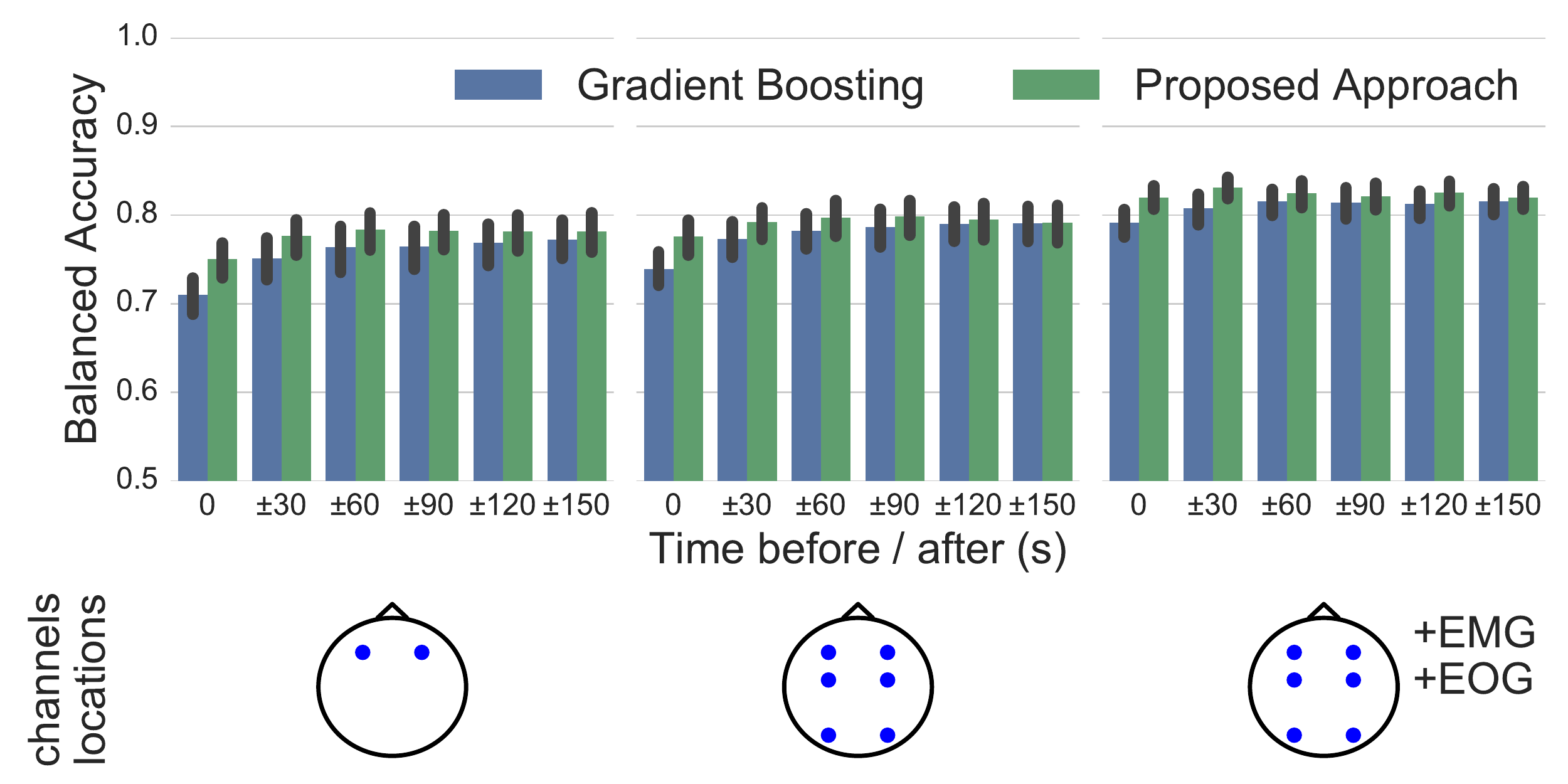}
\caption{\label{fig:influence_of_temporal_context}Influence of temporal context: considering the close temporal context induces a boost in performance especially when the spatial context is limited. From left to right: spatial configuration with $2$ frontal EEG channels, $6$ EEG channels, $6$ EEG channels plus $2$ EOG and $3$ EMG channels.}
\end{figure}

We considered the spatial configurations with $2$ frontal EEG channels, $6$ EEG channels, and $6$ EEG channels plus $2$ EOG and $3$ EMG channels. We varied the size of the temporal input sequence $S_k$ from $k=0$, \textit{i.e.} without temporal context, up to $k=5$. The classification results are reported in Fig.~\ref{fig:influence_of_temporal_context}.
%\agtodo{check this paragraph and say in caption of figure 8 what sensor configuration it corresponds to.}

\begin{figure*}[ht!]
\centering
\includegraphics[width=0.98\linewidth]{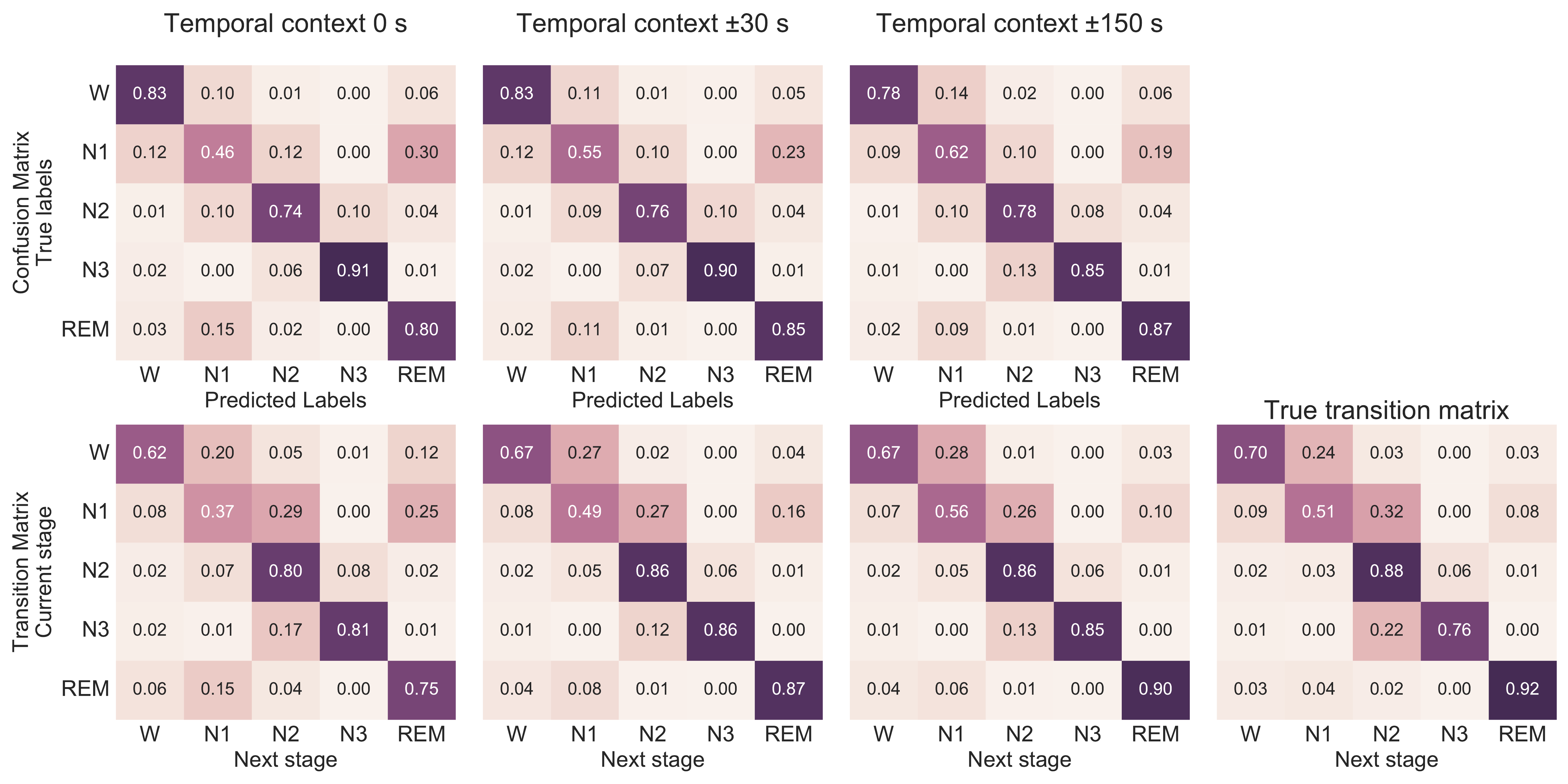}
\caption{\label{fig:confusion_transition_matrices}Influence of temporal context on the confusion matrices (top row) and the transition matrices (bottom row). Including more temporal context induces an increase of performance in the discrimination of stages N1, N2 and REM whereas it induces a slight decrease in the discrimination of W and N3 when the temporal context is too wide. Including more temporal context smooths the hypnogram.}
\end{figure*}

\color{black} 
%\agtodo{Here this is weird. You mention above the use of transition probabilities but you define them below. You should define them when you first mention them so 2 paragraphs above.}
% XXX: done
We furthermore evaluated the spatial configuration with only $2$ frontal EEG channels for which we report the average confusion matrices as well as the average transition matrices of the predicted hypnograms. We additionally included the transition matrix of the true hypnogram according to the labels given by the sleep expert. The matrices are presented in Fig~\ref{fig:confusion_transition_matrices}.
\color{black}

% \subsubsection{Temporal vs Spatial context}
\color{black} 
We observe in Fig.~\ref{fig:influence_of_temporal_context} that considering the close temporal context induces a boost in classification performances whereas considering a too large temporal context induces a decrease in performance. \color{black} The gain strongly depends on the spatial context taken into account. Indeed, our model trained on $2$ frontal channels with $-30 / +30$\,s of context achieves similar performances than with the $6$ EEG channel montage without temporal context. On the other hand, when considering an extended spatial context, the gain due to the temporal context turns out to be limited, as the performances of our approach or \emph{Gradient Boosting} with the $6$ EEG channels + $2$ EOG and $3$ EMG channels suggest.

%This experiment shows that our model clearly benefits from the temporal context, which validates the use of the features extractor distributed in time. It furthermore offers some insights about the influence of the close temporal context in relation to the spatial context.

\color{black}  
The finer analysis operated on the confusion matrices and transition matrices indicates a trade-off when integrating some temporal context: integrating the close temporal context brings benefits in the detection of some sleep stages specifically (N1, N2, REM) but a too large temporal context has a negative effect on the detection of W and N3 as emphasized by Fig.~\ref{fig:confusion_transition_matrices}.

Besides, the transition matrices of predictions compared to the true transition matrix in Fig.~\ref{fig:confusion_transition_matrices} indicate that processing a larger temporal context smooths the hypnogram. This corresponds to an increase of the diagonal coefficient in the transition matrices. As a consequence, the transition probabilities from stages W, N1, N2 and REM are improved but on the other hand, the transition probabilities from N3 (especially from N3 to N3) are negatively impacted.
\color{black}

\subsection{Experiment 4: More training data boost performance}
\begin{figure}[ht!]
\centering
\includegraphics[width=0.98\linewidth]{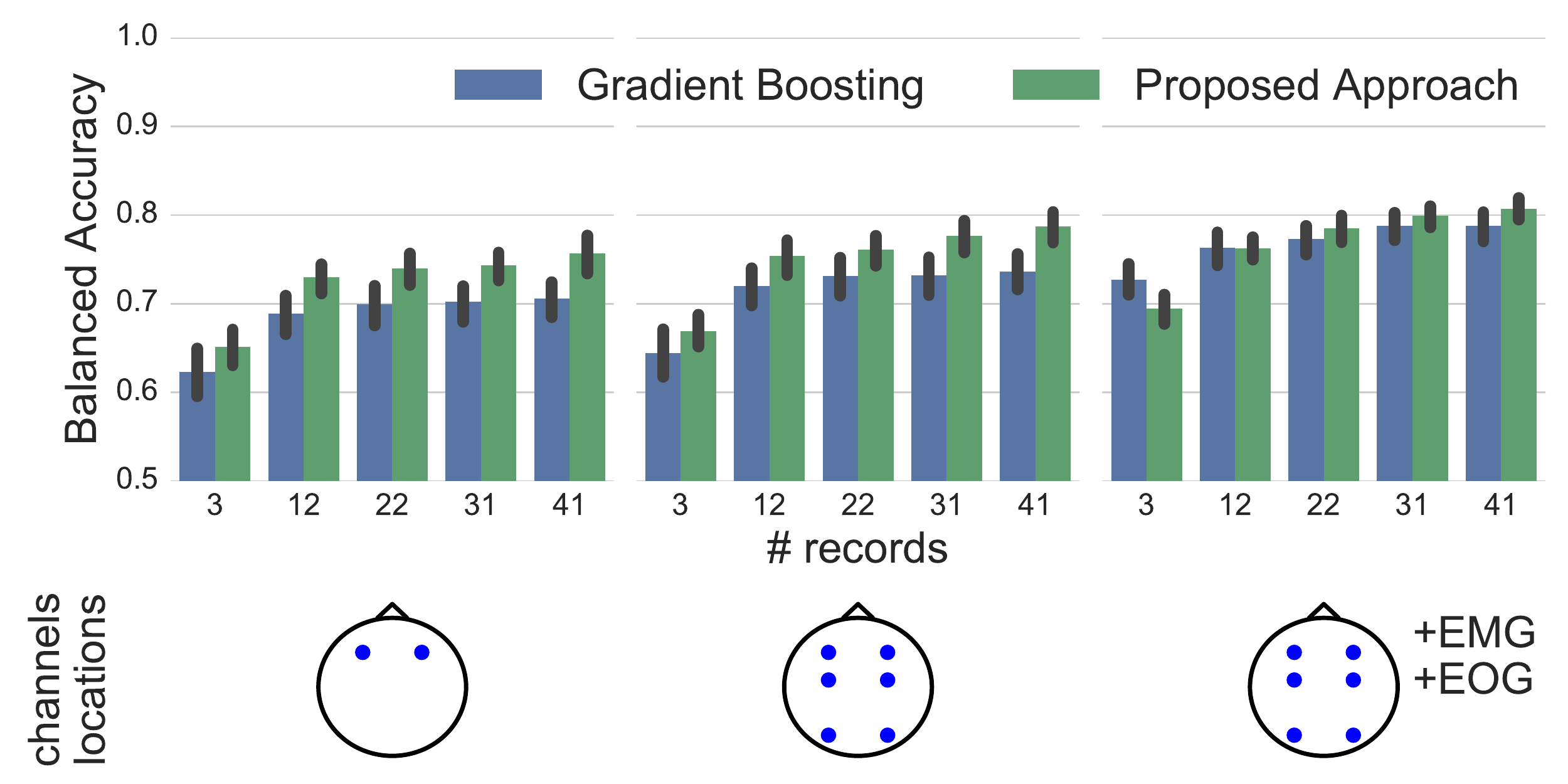} 
\caption{\label{fig:influence_of_training_set_size}Influence of the number of training records: the more training records the better performances are.
}
\end{figure}

In this experiment, we investigated the influence of the quantity of data on the classification performances of our approach.
% and demonstrate that its performances are directly linked to the quantity of training data at hand and the spatial context considered. Broadly speaking, an extended spatial context can partially cope with few training data and more training data can compensate for a small spatial context.
% 
% \subsubsection{Setup}
To do this we considered the spatial configurations with $2$ frontal EEG channels, $6$ EEG channels, and $6$ EEG channels plus $2$ EOG and $3$ EMG channels. Concretely, we varied the number of training records $n$ in $\left\{3, 12, 22, 31, 41\right\}$. We considered the same number of records for validation and testing as previously, \textit{i.e.} $10$. We furthermore carried out the experiments over $5$ random splits of training, validation and testing subjects. The classification results are reported in Fig.~\ref{fig:influence_of_training_set_size}.

% \ag{can you put 3 on the left and 41 on the right? it's more logical
% Ok done

% \subsubsection{The more data, the better performances}
\color{black} 
Every algorithm with any spatial context exhibits an increase in performance when there is more training data. \emph{Gradient Boosting} is more resilient than the proposed approach to the little data situation especially with a large spatial context. On the other hand, our deep learning model exhibits stronger increase in performance as a function of the quantity of data.
\color{black}

Furthermore, it appears that having few training records but an extended spatial context delivers as good performances as with many training records and few channels. Said differently, a rich spatial context can compensate for the scarcity of training data. Indeed, the input configuration with $6$ EEG channels plus $2$ EOG and $3$ EMG channels with only $12$ training subjects (right sub-figure) reaches the same performance as the $2$ EEG channels input configuration (left sub-figure) with $41$ training subjects.

\color{black} 
\subsection{Experiment 5: Opening the model box}
In this experiment, we aimed at understanding what the deep neural network learns. More precisely, we want to understand how the predictor relates a specific frequency content to the different sleep stages. We did so by occluding almost the whole frequency content, except a specific frequency band and monitoring the classification performances of the network while predicting on the filtered data. Such an operation, referred to as occlusion sensitivity has been successfully used to better understand how deep neural networks classify images~\cite{Zeiler2014}.

\begin{figure}[ht!]
\centering
\includegraphics[width=0.98\linewidth]{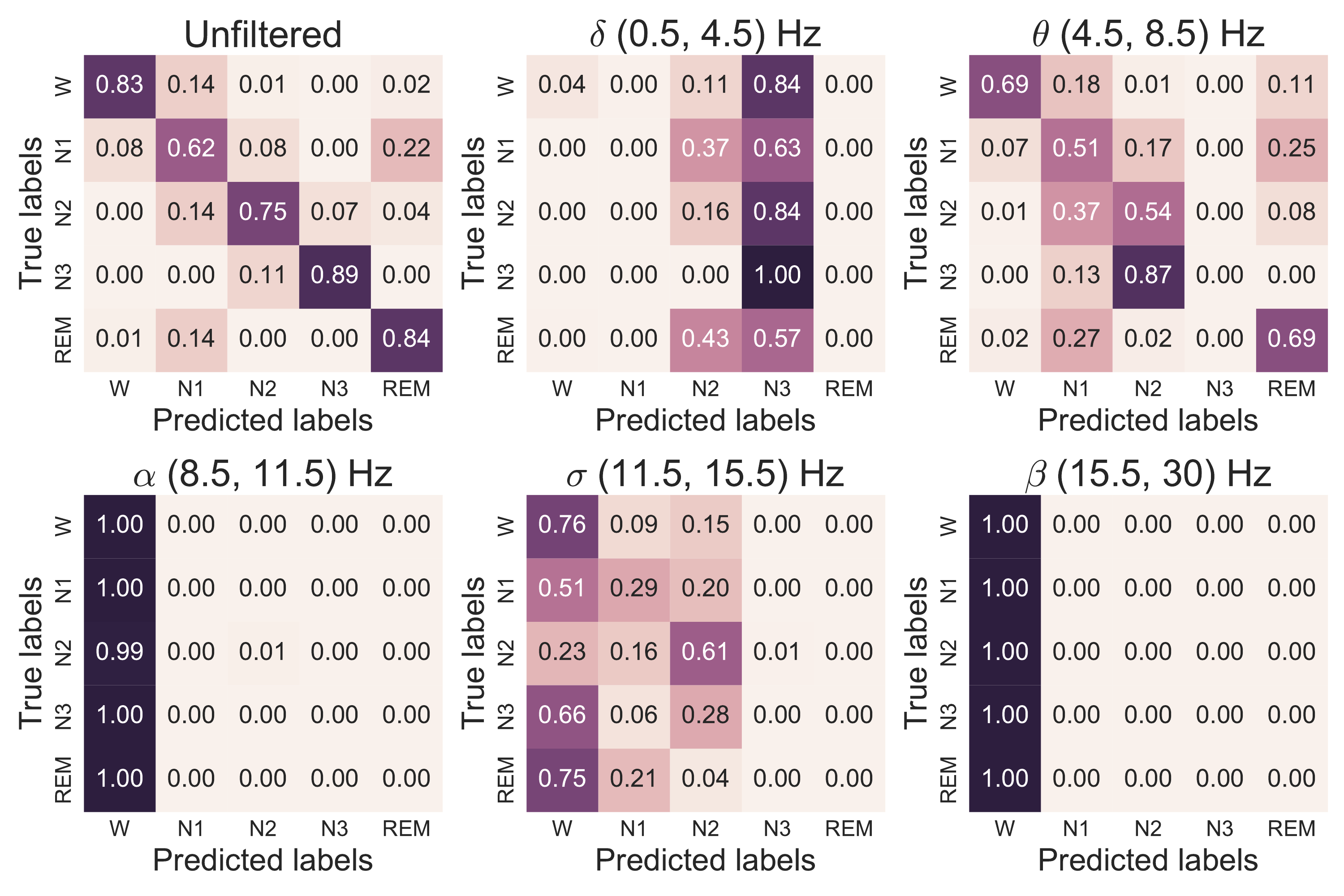}
\caption{\label{fig:prediction_on_filtered_data}Prediction on filtered data: confusion matrices associated to unfiltered and filtered signals from testing records.}
\end{figure}

We occluded almost the whole frequency domain and just kept a specific frequency band: either $\delta$ ($0.5 - 4.5 $ \,Hz), $\theta$ ($4.5 - 8.5$ \,Hz), $\alpha$ ($8.5-11.5$\,Hz), $\sigma$ ($11.5 - 15.5$\,Hz) or $\beta$ ($15.5 - 30$\,Hz). Each time, we took the neural network trained on the original signal, and made it predict on signals obtained after applying a band-pass filter with cutoff frequencies given by the considered frequency band. This means that for any filtered sample, the frequency content outside this frequency band was removed. We compared the predictions on the filtered signals with the original labels. The confusion matrices associated to the different band-pass filters are reported in Fig.~\ref{fig:prediction_on_filtered_data}.

Using the network on filtered signals enables to reveal the relationship between a specific frequency content and the sleep stages predicted by the network. Indeed, when only the delta band is kept, the network assigns N2 or N3 to all the samples. This implies that the network associates a low frequency content to N2 and N3 stages where there are actually low frequency events such as slow oscillations or K-complex.

Similarly, we observe that when the network predicts on signals where only the alpha band is kept, the network predicts mostly W. This is in agreement with the rules human scorers follow. A similar approach could be performed with much finer frequency bands.

Thus, despite the black-box nature of the proposed approach, this occluding procedure allows to open the box and to reveal interesting insights about how the model relates a particular frequency content to the different sleep stages.
\color{black}

\color{black} 
\section{Discussion}
\label{sec:5_discussion_conclusion}
In this section, we discuss the architecture characteristics of our approach and put them in perspective with state-of-the art methods. We furthermore discuss the use of temporal context to take into account transitions between sleep stage and discuss its use for applications. Finally, we discuss points about the training of the proposed architecture and how this one can meet a specific need.

\subsection{Spatial filtering}
The proposed architecture was designed to handle a multivariate input thanks to a spatial filtering step. This step is motivated by the fact that a linear combination of the input channels should enhance the information useful for the task, and so even more if the spatial filters are optimized via back propagation on the training data. Motivated by simplicity, we chose the number of virtual channels equal to the number of input channels. Yet, this constitutes a degree of freedom one may play with to increase the performances of the network as was explored in~\cite{Cecotti2011}.

As a comparison, \cite{Biswal2017} averages the input time series to obtain a single one which is then fed into a 1D convolutional network. This can be seen as a particular case of our spatial filtering step where the number of virtual channels is equal to $1$ and where the unique spatial filter coefficients are fixed to $1 / C$, with $C$ the number of input channels. On the contrary, \cite{Stephansen2017} proposed an approach that also takes as input a multivariate time series but does not perform a particular spatial processing.

\subsection{Feature extractor architecture}
The proposed feature extractor exhibits a simple and versatile $2$ layer architecture. Considering fewer or more layers was explored but did not deliver any extra gain in performance. We furthermore opted to perform spatial and temporal convolutions strictly separately. By doing so we replaced possible $2$D expensive convolutions by a $1$D spatial convolution and a $1$D temporal convolution. Such a low rank spatio-temporal convolution strategy turned out to be successful in our experiments.

Regarding the dimensions of the convolution filters and pooling regions, our approach was motivated by the ability of neural networks to learn a hierarchical representation of input data, extracting low level and small scale patterns in the shallow layers and more complex and large scale patterns in the deep layers. Our strategy is quite different from~\cite{Tsinalis2016, Supratak2017} which use large temporal convolution filters. Despite the use of smaller filters, Fig.~\ref{fig:benchmark_general_metrics} and Fig.~\ref{fig:prediction_on_filtered_data} demonstrate that our architecture is able to discriminate stages with low frequency content, such as $N3$, from stages with higher frequency content such as $N2$ due to the presence of spindles, or even from $W$ and $N1$ with the presence of $\alpha$ ($8-12$Hz) bursts. Besides, our proposed architecture turns out to be data agnostic and handles well both EEG, EOG and EMG signals as shown by the results of experiment 2, see Fig.~\ref{fig:influence_spatial_configuration} and Fig~\ref{fig:influence_additional_modalities}.

Yet it is to be noticed, that recent approaches use even smaller convolution filters, of size $2$, $3$, $5$, or $7$ ~\cite{Stephansen2017, Biswal2017, Sors2017}. On the contrary they also use a larger number of features maps from $64$ up to $512$~\cite{Stephansen2017, Sors2017}. The use of small filters in combination with a larger number of features maps is worth investigating and quantifying and might result in more signal agnostic neural networks.

\subsection{Number of parameters}
The complexity of the proposed network and its number of parameters are quite small thanks to specific architecture choices. The overall network does not exhibit more than $\sim 10^4$ parameters when considering an extended spatial context, and not more than $\sim 10^5$ parameters when considering both an extended spatial context and an extended temporal context. This is quite simple and compact compared to the recent approaches in~\cite{Tsinalis2016} which has up to $\sim 14.10^7$ parameters and \cite{Supratak2017} which exhibits $\sim 6.10^5$ parameters for the feature extractor and $2. 10^7$ parameters for the sequence learning part using BiLSTM. This significant difference with~\cite{Tsinalis2016} is mainly due to our choice of using small convolution filters ($64$ time steps after low pass filtering and downsampling), large pooling regions (pooling over $16$ time steps) according to the $128$\,Hz sampling frequency and removing the penultimate fully connected layers before the final softmax classifier. Such a strategy has already been successul in computer vision~\cite{Springenberg2015} and EEG~\cite{Lawhern2016}. %Regarding the comparison with \cite{Supratak2017}, our approach is different in the way that it builds a features representation as explained previously but also in the way how it processes the temporal context without using any recurrent neural network.

\subsection{Classification metrics}
The proposed approach yields equal (univariate) or higher (multivariate) classification metrics than the other benchmarked feature extractors while presenting a limited training run time per epoch or prediction time per night record (cf. Fig.~\ref{fig:benchmark_general_metrics}). The analysis of per class metrics shows that the proposed approach might not reach the highest performance on every stages (cf. Fig.~\ref{fig:benchmark_per_class_metrics}). Indeed, \emph{Supratak et al. 2017} outperforms on N1, and \emph{Gradient  Boosting} exhibits a similar accuracy in N3. However, the proposed approach performs globally well and appears to be quite robust in comparison to the other approaches.

The proposed approach is particularly good at detecting W (high sensitivity $0.85$ and specificity close to $1$). This characteristic might be particularly interesting for clinical applications where a diagnosis of fragmented sleep might rely on the detection of W.

In order to measure the relevance of our approach for different types of subjects, we monitored the balanced accuracy of a subject as a function of the sleep fragmentation index (total number of awakenings and sleep stage shifts divided by total sleep time)~\cite{Haba-Rubio2004}. The results (not shown) did not exhibit a particular correlation between this measure of sleep quality and the classification performances. This indicates that the proposed approach could be used for clinical purposes with patients whose sleep exhibit abnormal structures.

Unfortunately, the different classification performances cannot be compared with inter-scorer agreement on this dataset since the night records have only been annotated by a single expert. Yet, a $0.80$ agreement has been reported between scorers~\cite{Rosenberg2014}. Furthermore, \cite{Stephansen2017} monitored the classification accuracy of their model as a function of the consensus from 1 to 6 scorers. The reported curve was linearly increasing from $0.76$ accuracy for 1 scorer up to 0.87 accuracy for a 6 scorer consensus. We shall reproduce such an experiment with the proposed approach in our future work.

\subsection{Temporal context and transitions}
Our architecture allows naturally  to learn from the temporal context as it only relies on the aggregation of temporal features and a softmax classifier. Such a choice, enabled us to measure the influence of the close temporal context and better understand its impact. It differs from the approaches proposed by \cite{Tsinalis2016, Sors2017} as our features extractor always receives $30$\,s of signals, and is therefore applied to a sequence of neighboring $30$\,s samples. On the contrary, \cite{Tsinalis2016, Sors2017} extended the feature extractor input window to $150$\,s, respectively $120$\,s. In \cite{Supratak2017}, a temporal context of $25$ neighboring $30$\,s samples is processed.

Our experiment on temporal context highlights a trade-off when integrating some temporal context: integrating some temporal context brings benefits in the detection of some sleep stages specifically (N1, N2, REM) but a too large temporal context has a negative effect on the detection of W and N3 stages as emphasized by Fig.~\ref{fig:confusion_transition_matrices}. This naturally translates to the balanced accuracy scores which exhibit a significative increase for small temporal context and no increase, or even a decrease, for large temporal context (cf. Fig.~\ref{fig:influence_of_temporal_context}). Looking at the transitions matrices, it appears that more temporal context smoothes the hypnograms which might be detrimental to the quality of the system. For these reasons, temporal context should be used, but its width must be cross-validated.

Besides, some subjects might exhibit abnormal sleep structures related to a sleep disorder~\cite{Rosenberg2014}. There is thus a trade-off between boosting the classification performance by integrating as much context as possible and not over-fitting sleep transitions in order to not miss a sleep disorder related to a fragmented sleep. This is an additional argument in favor of cross-validating the temporal context width.

An extension of our approach, for example  to capture complex stage transitions or long term dependencies would be to employ a recurrent network architecture. Along these lines recent approaches have proposed more complex strategies to integrate the temporal context with LSTM unit cells or Bi-LSTM unit cells~\cite{Stephansen2017, Biswal2017, Supratak2017, Dong2016, Hochreiter1997}. Integrating our feature extractor with such recurrent networks remains to be done and should lead to further performance improvements.

\subsection{Influence of dataset}
Figure~\ref{fig:influence_of_training_set_size} raises an important question: how much data is needed to establish a correct benchmark of predictive models for sleep stage classification? This is particularly interesting concerning the deep learning approaches. Indeed, the \emph{Gradient Boosting} handles quite well the small data situation and does not exhibit a huge increase in performances with the increase of the number of training records. On the contrary our approach delivers particularly good performances if enough training data are available. Extrapolation of the learning curves (performance as a function of the number of training records) in Fig.~\ref{fig:influence_of_training_set_size} suggests that one could expect better performances if more data were accessible. This forces us to reconsider the way we compare predictive models when training dataset sizes differ between experiments since the quantity of training data plays the role of a hyper-parameter for some algorithms like ours. Some algorithms become indeed better when more data are available (see for example Fig.~1 in \cite{Banko:2001}).

\subsection{Choice of sampling and metrics}
Our approach was particularly motivated by the accurate detection of any sleep stage independently to its proportion. To achieve this goal, all approaches have been trained using balanced sampling and evaluated with balanced metrics (except for experiment 1 where more metrics have been used). We observed that the choice of sampling strategies employed during online learning impacts the evaluation metrics and conversely the choice of metrics should motivate the choice of sampling strategies. Indeed, balanced sampling should be used to optimize the balanced accuracy of the model. On the other hand, random sampling should be used to boost the accuracy. The use of balanced sampling has been reportedly used or commented in previous works~\cite{Tsinalis2016, Supratak2017, Sors2017}.

Nonetheless, for a specific clinical application, one may decide that errors on a minor stage, such as $N1$, are not so dramatic and hence prefer to train the network with random batches of data. On the contrary, one might want to discriminate as accurately as possible $N1$ stages from $W$ or $REM$ and therefore one should use balanced sampling, or over sampling of N1.

Sampling strategy and evaluation metrics is a degree of freedom one can play with to adapt the network for his own experimental or clinical purposes.

\color{black}

\section{Conclusion}
\color{black}
In this study we introduced a deep neural network to perform temporal sleep stage classification from multimodal and multivariate time series. The model pools information from different sensors thanks to a linear spatial filtering operation and builds a hierarchical features representation of PSG data thanks to temporal convolutions. It additionally pools information from different modalities processed with separate pipelines.

The proposed approach in this paper exhibits strong classification performances compared to the state-of-the-art with a little run time and computational cost. This makes the approach a potential good candidate for being used in a portable device and performing online sleep stage classification.

Our approach enables to quantify the use of multiple EEG channels and additional modalities such as EOG and EMG. Interestingly, it appears that a limited number of EEG channels (6 EEG: F3, F4, C3, C4, O1, O2) gives performances similar to 20 EEG channels. Furthermore, using EMG channels boosts the model performances.

The use of temporal context is analyzed and quantified and appears to give significant increase in performance when the spatial context is limited. It is to be noticed that the temporal context as explored in this paper might not be directly suitable for online prediction, but it is easily usable for offline prediction.
\color{black}

{
\bibliographystyle{ieeetr}
\bibliography{biblio}

}

\end{document}